\documentclass[runningheads]{llncs}

\usepackage[mobile]{iciap}

\usepackage{iciapabbrv}

\usepackage{wrapfig}
\usepackage{graphicx}
\usepackage{placeins}

\usepackage[accsupp]{axessibility}  %

\usepackage{enumitem}
\usepackage{booktabs}  
\usepackage{multirow}  
\usepackage{makecell}  
\usepackage[table]{xcolor}
\usepackage{bm}
\usepackage{algorithm}
\usepackage[noend]{algpseudocode}
\algnewcommand\algorithmicinput{\textbf{Input:}}
\algnewcommand\algorithmicoutput{\textbf{Output:}}
\algnewcommand\Input{\item[\algorithmicinput]}
\algnewcommand\Output{\item[\algorithmicoutput]}

\usepackage{hyperref}

\usepackage{orcidlink}

\begin{document}

\title{Towards Robust Knowledge Removal in Federated Learning with High Data Heterogeneity}

\titlerunning{Robust Knowledge Removal in Federated Learning }

\author{Riccardo Santi \orcidlink{0009-0007-9146-2401} \and Riccardo Salami \orcidlink{0009-0002-0704-5810} \and
Simone Calderara \orcidlink{0000-0001-9056-1538}}

\authorrunning{R. Santi et al.}

\institute{AImageLab, University of Modena and Reggio Emilia, Italy\\
\email{\{riccardo.santi, riccardo.salami, simone.calderara\}@unimore.it}
}

\maketitle

\begin{abstract} 
Nowadays, there are an abundance of portable devices capable of collecting large amounts of data and with decent computational power. This opened the possibility to train AI models in a distributed manner, preserving the participating clients' privacy. However, because of privacy regulations and safety requirements, elimination upon necessity of a client contribution to the model has become mandatory. The cleansing process must satisfy specific efficacy and time requirements. In recent years, research efforts have produced several knowledge removal methods, but these require multiple communication rounds between the data holders and the process coordinator. This can cause the unavailability of an effective model up to the end of the removal process, which can result in a disservice to the system users. In this paper, we introduce an innovative solution based on Task Arithmetic and the Neural Tangent Kernel, to rapidly remove a client's influence from a model.
\end{abstract}

\section{Introduction}
In recent years \textbf{Federated Learning} (FL)~\cite{fl1} has emerged as a way to train \textbf{Deep Learning} models in a distributed and privacy-preserving manner. In certain cases, it becomes essential to remove the knowledge contributed by a group of clients during the FL process, while preserving the model’s performance. This need arises, for instance, when a client’s data is found to be poisoned—leading the model to exhibit malicious behavior~\cite{backdoor1} — or when compliance with privacy regulations such as GDPR~\cite{gdpr} and CCPA~\cite{ccpa} is required, or in sensitive domains such as medical settings~\cite{2024IEEEACCESS,bontempo2023graph} where data revocation may be necessary to preserve patient confidentiality. In these cases, it is crucial to maintain a performing global model even after removing some of the client's knowledge.

\noindent
\textbf{Federated Unlearning} (FU)~\cite{fu} algorithms are able to address this necessity by removing the traces of the data of the target client, i.e. the client that needs to be unlearned, without compromising the privacy guarantees required by the decentralized learning process. Recent studies have introduced various FU methods that produce usable models, while minimizing unlearning time~\cite{fedrecover, fedrecovery}. The best-known method is \textbf{FedEraser}~\cite{federaser}, which recalibrates and redirects past model updates to estimate the global model as if it had never been influenced by the client to be unlearned. The downside of this method, just like many other FU algorithm, is that it needs to execute several communication rounds before obtaining a new effective model. The unavailability of an effective model during the execution of an unlearning algorithm, can disrupt service, particularly in systems serving a significant number of users.
Our contributions are as follows.
\begin{itemize}[noitemsep, topsep=0pt, leftmargin=*]
    \item We present a novel FU method that allows to unlearn a client contribution with a single communication round, minimizing the time required to obtain a functional model.
    \item We evaluate its performance against a baseline and two competitors through extensive experiments on two different datasets.
\end{itemize} 

\section{Related Works}
\textbf{Federated Learning.}\xspace FL is a setting designed to train a model when the training data is scattered across multiple devices that cannot share them to preserve privacy. The training procedure is divided into communication rounds and is orchestrated by a \textbf{central server} $\mathcal{S}$~\cite{lorm, salami2024federated}. 

\noindent
The baseline FL algorithm is Federated Averaging (FedAvg)~\cite{fedavg}, which aggregates client updates, computed on their local data, via simple parameter averaging. To address heterogeneity, FedProx~\cite{fedprox} introduces a proximal term that keeps local updates close to the global model, while SCAFFOLD~\cite{karimireddy2020scaffold} adds control variates to mitigate client drift. FedDC~\cite{gao2022feddc} lets clients adjust updates by estimating and correcting local drifts before aggregation. GradMA~\cite{luo2023gradma} projects gradients into a compact memory space, balancing local optimization with alignment to the server model. Fisher-Weighted Averaging~\cite{matena2022merging} further modifies FedAvg by weighing client parameters using their Fisher information. Finally, FedProto~\cite{tan2022fedproto} uses class prototypes, computed client-side and averaged by the server, for prototype-based aggregation.

\noindent
\textbf{Machine Unlearning}~\cite{mu1} was introduced with the purpose of performing unlearning on centralized Machine Learning models. However, these techniques result futile due to the decentralized nature of FL.
\noindent
To address this limitation, the concept of Federated Unlearning has emerged. The most known FU algorithm is FedEraser~\cite{federaser}, which recalibrates past model updates estimating the parameters of a cleansed model. FedRecover~\cite{fedrecover} addresses poisoning attacks by recovering a clean model using historical client updates and Hessian approximations to avoid costly retraining. FedRecovery~\cite{fedrecovery} instead proposes a retraining-based method without storing any historical information, leveraging diagonal Fisher Information and momentum-based updates to achieve unlearning efficiently. Ferrari~\cite{gu2024ferrari} introduces a new metric to evaluate the sensitivity of the model to the data features. Unlearning is achieved by desensitizing the model w.r.t. the target features.  

\section{Methodology}\label{method}

\subsection{Background}
\paragraph{\textbf{Task arithmetic}}~The term Task Arithmetic (TA)~\cite{ta, rinaldiupdate, secondOrder} denotes a framework for the editing of pre-trained deep learning models through simple linear operations in the parameter space. Given a pre-trained model, we indicate its parameters with $\theta_0 \in \mathrm{R}^d$. Once the model is fine-tuned on a task $t$, we indicate its parameters as $\theta_t \in \mathrm{R}^d$, and we refer to the difference between the pre-trained and fine-tuned parameters $\tau_t = \theta_t - \theta_0 $ as \textbf{Task Vector}. Task vectors embed how a model changes when fine-tuned on a specific task, and if combined, they allow the model to generalize across different tasks or remove learned information, without the need to retrain the model. Given a set of $T$ task vectors $\left\{ \tau_1, \tau_2, \dots, \tau_T  \right\}$, the updated model is equal to $\theta_{new} = \theta_0 + \sum_{i=1}^T\lambda_i\tau_i,$ where $\lambda_i$ are scalar coefficients that weigh the relevance of each task vector.

\paragraph{\textbf{Neural Tangent Kernel}}~The Neural Tangent Kernel (NTK)~\cite{ntk} is a kernel that, in the infinite-width limit, linearizes the learning dynamics of a neural network. As the width of a neural network increases, the update of its parameters becomes infinitesimal, allowing us to approximate the model changes during training with a first-order Taylor expansion $f_{lin}(x; \theta) = f(x; \theta_0) + (\theta - \theta_0)^\top \nabla_\theta f(x; \theta_0).$
Recent studies have demonstrated that training in the NTK regime enhances a model’s task arithmetic~\cite{tantk}. This improvement stems from increased \textbf{disentanglement} of the model parameters, reflecting how little task vectors interfere with one another, potentially degrading the aggregated model’s performance. Higher weight disentanglement leads to reduced task interference, thus superior task arithmetic.
So, given a set of $T$ tasks on which we fine-tune $T$ different models $f_{lin}$ all sharing the same pre-training $\theta_0$, the final aggregated model is:
\begin{equation}
    f_{lin}\left(x;\theta_0 + \sum_{t\in T} \lambda_t \tau_t\right) = f\left(x;\theta_0\right) + \sum_{t\in T}\left( \lambda_t \tau_t\right)^\top \nabla_\theta f\left(x;\theta_0\right)
\end{equation}

\subsection{Method}

We present our novel FU method, which, unlike most existing approaches, delivers a cleansed model in just one communication round. Additional rounds can then be performed, as with other methods, to further unlearn the target client and improve performance on the remaining clients.

\noindent
At the beginning of the FL procedure the clients are provided with the pre-trained weights $\theta_0$ of the server model. Starting from this pre-train, they will fine-tune in parallel two task vectors independent of each other on the local dataset $\mathcal{D}_k$. The first task vector $\tau_k$ will be involved in the distributed training process and will contribute to the computation of the global model. On the contrary, the second task vector, which we call \textbf{standalone task vector} $\tau^{sa}_k$, is kept isolated so as to keep clean of the information obtained by other clients, which may eventually need to be unlearned in the future.

\noindent
When the unlearning phase begins, the target client $tgt$,  transmits its $\tau^{sa}_{tgt}$. According to task arithmetic, these parameters are a task-specific fine-tuning so, to unlearn the client $tgt$, we can simply subtract $\tau^{sa}_{tgt}$ from the global model:

\begin{equation}
    \hat{\theta}_{clean} = \hat{\theta} - \lambda_{tgt}\tau_{tgt}^{sa},
\end{equation}

\noindent
where $\hat{\theta}_{clean}$ is the new global model free of the target client influence, $\hat{\theta}$ is the global model prior to the unlearning process, and $\lambda_{tgt}$ is a scalar weight indicating the relevance of that task vector of the client we need to unlearn.

\noindent
To further enhance the task arithmetic of the model, on each client $k$ both the main task vector $\tau_k$ and the standalone $\tau_k^{sa}$ are trained in the NTK regime. This allows us to model the new global model as follows:

\begin{align}\label{eq_method}
    f_{lin}\Big(x;\theta_{r-1} &+ \sum_{k=1}^K \lambda_k \tau_k - \lambda_{tgt} \tau_{tgt}^{sa}\Big) \nonumber\\
    &= f\left(x;\theta_{r-1}\right) + 
    \left( \sum_{k=1}^K \lambda_k \tau_k 
    - \lambda_{tgt} \tau_{tgt}^{sa} \right)^\top 
    \nabla_\theta f\left(x;\theta_{r-1}\right),
\end{align}

\noindent
where $K$ is the number of clients involved in the FL process, $tgt$ is the target client, $\theta_{r-1}$ are the model weights at second-to-last FL round, and $ \sum_{k=1}^K \lambda_k \tau_k$ are the client contributions right before the beginning of the FU algorithm. This ensures that only the knowledge of the $tgt$ client is removed from the server, while preserving the knowledge contributed by the other clients.
We refer to this method as \textit{Stand Alone TA} (SATA).

\section{Experiments}

Each of our experiments consists of three phases:
\begin{itemize}[noitemsep, topsep=0pt, leftmargin=*]
    \item \textit{Federated Learning} (FL) phase, in which we train a global model according to the FedAvg algorithm.
    \item \textit{Federated Unlearning} (FU) phase, where we execute a federated unlearning algorithm to remove the influence of the first client.
    \item \textit{Post Unlearning} (PU) phase, in which we resume the federated learning procedure, without the first client.
\end{itemize}

\noindent
We fine-tune the ViT-B/16 from OpenAI’s CLIP model~\cite{clip}. The features extracted are passed to a classification head consisting of a single dense layer initialized with the average embeddings of class-specific textual templates drawn from the dataset. The head is kept frozen.

\subsubsection{Benchmarking} We evaluate our method on \textbf{Cars-196}~\cite{krause20133d} --  a dataset of car images grouped in 196 classes corresponding to vehicle brand, model, and year -- and \textbf{Resisc45}~\cite{cheng2017remote}, a set of remote sensing images grouped into 45 classes. As in standard FL practice, each dataset is partitioned using a Dirichlet distribution to mimic real-world conditions where data is distributed in a non-I.I.D. fashion. The parameter $\beta$ controls the degree of skewness: the smaller its value, the more skewed the data distribution.
We evaluate the global model by measuring its accuracy at the end of each communication round. Unlearning performance is assessed by tracking the model’s accuracy on the target client’s testing data after each round, starting from the first round where unlearning is applied. To compare with state-of-the-art methods, we tested our approach against the following FU techniques:
\begin{itemize}[noitemsep, topsep=0pt, leftmargin=*]
    \item \textit{Train From Scratch} (TFS) (upper bound): the model is retrained starting from the pre-training without the target client.
    \item \textit{Continue to Train} (CTT)~\cite{nguyen2024empirical}: unlearning is achieved only with the exclusion of the client which needs to be unlearned, leveraging the phenomenon of catastrophic forgetting~\cite{clcatastrophicforgetting,menabue2024semantic,frascaroli2024clip}. 
    \item \textit{FedEraser}~\cite{federaser}: it reconstructs the global model by leveraging historical client updates stored on the central server. Starting from the initial model weights, the clients perform calibration epochs, and the information obtained is used to recalibrate the stored updates.
\end{itemize}

\begin{table*}[t]\renewcommand{\arraystretch}{1.3}
\caption{Accuracy ($\downarrow$) on the testing set of the unlearned client at the end of the first FU round (FU), and at the end of the whole pipeline (PU).}
\label{model_target_acc}
\centering
\setlength{\tabcolsep}{0.4em}
\resizebox{\textwidth}{!}{%
\begin{tabular}{l@{\hskip 15pt}|*{2}{c}|*{2}{c}|*{2}{c}|*{2}{c}|*{2}{c}|*{2}{c}}
\hline
\multirow{3}{*}{Method}         & \multicolumn{4}{c}{$\beta$ = 0.05} & \multicolumn{4}{c}{$\beta$ = 0.1} & \multicolumn{4}{c}{$\beta$ = 0.5} \\ \cline{2-13}
                                & \multicolumn{2}{c}{Resisc45} & \multicolumn{2}{c}{Cars196} & \multicolumn{2}{c}{Resisc45} & \multicolumn{2}{c}{Cars196} & \multicolumn{2}{c}{Resisc45} & \multicolumn{2}{c}{Cars196} \\ \cline{2-13} 
                                &  FU  & PU &  FU &  PU  &  FU  & PU &  FU &  PU  &  FU  & PU &  FU &  PU  \\ \hline

TFS             & 59.61 & 65.52    & 51.01 & 47.51      & 88.00 & 92.72       & 55.75 & 54.83    	& 93.20 & 94.71     & 80.07 & 81.65    \\ \hline

CTT             & 70.90 & 66.84    & 74.79 & 60.95      & 92.62 & 90.05       & 75.09 & 65.12    	& 93.27 & 94.03     & 85.76 & 85.98    \\ \hline

FedEraser       & 56.79 & 62.96    & 56.04 & 51.18      & 80.10 & 87.18       & 58.32 & \textbf{56.61}    	& 89.95 & 92.59     & 69.93 & 80.61    \\ \hline

SATA NTK      & \textbf{9.96} & \textbf{53.70}               & \textbf{1.48} & \textbf{49.17}           & \textbf{31.69} & \textbf{81.44}                 & \textbf{6.36} & 57.59   	      & \textbf{14.29} & \textbf{90.48}                  & \textbf{0.27} & \textbf{74.15}    \\ \hline

\end{tabular}
}
\end{table*}

\begin{table*}[t]\renewcommand{\arraystretch}{1.5}
\caption{Accuracy ($\uparrow$) at the end of the last FL round (FL), at the end of the first FU round (FU), and at the end of the whole pipeline (PU).}
\label{model_acc}
\centering
\setlength{\tabcolsep}{0.4em}
\resizebox{\textwidth}{!}{%
\begin{tabular}{l@{\hskip 10pt}|*{3}{c}|*{3}{c}|*{3}{c}|*{3}{c}|*{3}{c}|*{3}{c}} 
\hline
\multirow{3}{*}{Method}         & \multicolumn{6}{c}{$\beta$ = 0.05} & \multicolumn{6}{c}{$\beta$ = 0.1} & \multicolumn{6}{c}{$\beta$ = 0.5} \\ \cline{2-19}
                                & \multicolumn{3}{c}{Resisc45} & \multicolumn{3}{c}{Cars196} & \multicolumn{3}{c}{Resisc45} & \multicolumn{3}{c}{Cars196} & \multicolumn{3}{c}{Resisc45} & \multicolumn{3}{c}{Cars196}  \\ \cline{2-19}
                                & FL  & FU &  PU &  FL  & FU &  PU  & FL  & FU &  PU &  FL  & FU &  PU & FL  & FU &  PU &  FL  & FU &  PU\\ \hline

TFS                 & 90.69 & 91.56 & 93.79    & 85.77 & 86.54 & 88.54       & 93.18 & 94.07 & 95.14    & 86.05 & 86.25 & 87.73      & 96.04 & 95.98 & 96.6	      & 87.03 & 87.37 & 89.06 \\ \hline

CTT                 & \textbf{89.33} & \textbf{91.97} & \textbf{94.23}    & \textbf{83.60} & \textbf{85.62} & \textbf{89.12}       & \textbf{93.22} & \textbf{93.24} & \textbf{95.19}    & \textbf{85.61} & \textbf{86.42} & \textbf{88.73}      & \textbf{96.05} & \textbf{95.78} & \textbf{96.42}	  & \textbf{87.79} & \textbf{88.16} & \textbf{89.81} \\ \hline

FedEraser           & 89.40 & 84.30 & 91.56    & 84.77 & 73.11 & 86.18       & 93.21 & 83.62 & 93.48    & \textbf{85.61} & 75.07 & 85.47      & 95.75 & 93.73 & 95.68	  & 87.74 & 75.35 & 87.61 \\ \hline

SATA NTK            & 84.43 & 73.71 & 90.24    & 81.78 & 20.00 & 85.86       & 88.41 & 76.92 & 91.85    & 83.26 & 44.47 & 86.98      & 93.48 & 53.12 & 94.96	  & 85.61 & 2.53 & 86.31 \\ \hline

\end{tabular}
}
\end{table*}

\noindent
For Resisc45, all experiments include 3 communication rounds with 3 training epochs each, except FedEraser’s 1-epoch calibration round and our method, which unlearns clients in a single round without extra training. For this reason, we increase the number of PU rounds so that the total number of rounds matches the rounds executed by the competitors.
For Cars-196, the FL phase runs for 10 rounds, followed by 10 FU rounds and 5 post-unlearning rounds, each with 3 training epochs. Again, FedEraser uses 1-epoch calibration rounds, while our method uses a single communication round to unlearn and extends PU rounds to match the competitors' total rounds.
Clients employ the AdamW optimizer~\cite{loshchilov2019decoupled}. The results reported in \cref{model_acc}, \cref{model_target_acc}, \cref{model_acc_ablation}, and \cref{model_target_acc_ablation} have been obtained with a grid search of $\lambda_{tgt}$ of \cref{eq_method} and the learning rate for both $\tau_{k}$ and $\tau_{k}^{sa}$  for each $\beta$ used.
All method-dataset pairs are evaluated using 3 different Dirichlet coefficients.

\begin{table*}[t]\renewcommand{\arraystretch}{1.5}
\caption{Accuracy ($\uparrow$) at the end of: the last FL round (FL), the first FU round (FU), and the whole pipeline (PU).}
\label{model_acc_ablation}
\centering
\setlength{\tabcolsep}{0.4em}
\resizebox{\textwidth}{!}{%
\begin{tabular}{l@{\hskip 10pt}|*{3}{c}|*{3}{c}|*{3}{c}|*{3}{c}|*{3}{c}|*{3}{c}} 
\hline
\multirow{3}{*}{Method}         & \multicolumn{6}{c}{$\beta$ = 0.05} & \multicolumn{6}{c}{$\beta$ = 0.1} & \multicolumn{6}{c}{$\beta$ = 0.5} \\ \cline{2-19}
                                & \multicolumn{3}{c}{Resisc45} & \multicolumn{3}{c}{Cars196} & \multicolumn{3}{c}{Resisc45} & \multicolumn{3}{c}{Cars196} & \multicolumn{3}{c}{Resisc45} & \multicolumn{3}{c}{Cars196}  \\ \cline{2-19}
                                & FL  & FU &  PU &  FL  & FU &  PU  & FL  & FU &  PU &  FL  & FU &  PU & FL  & FU &  PU &  FL  & FU &  PU\\ \hline

$\theta_0 + \sum_{i \neq tgt}\lambda_i \tau_i^{sa}$                & 88.67 & 84.67 & \textbf{93.94}    & \textbf{84.82} & \textbf{75.23} & \textbf{86.82}       & \textbf{93.25} & \textbf{85.05} & \textbf{94.54}    & \textbf{85.92} & 75.14 & \textbf{87.44}      & 95.78 & \textbf{94.43} & \textbf{96.32}	  & \textbf{87.95} & \textbf{80.85} & \textbf{88.50} \\ \hline

$\theta_0 + \sum_{i \neq tgt}\lambda_i \tau_i^{sa}$ NTK            & 85.00 & 84.75 & 91.83    & 82.33 & 74.24 & 85.37       & 89.90 & 83.46 & 92.39    & 83.83 & \textbf{76.81} & 86.55      & 92.10 & 90.98 & 94.21	  & 85.31 & 79.12 & 87.19 \\ \hline

$\hat{\theta} - \lambda_{tgt}\tau_{tgt}^{sa}$                & \textbf{88.90} & \textbf{89.47} & 93.11    & 84.55 & 52.10 & 86.66       & 92.59 & 78.57 & 93.99    & 84.78 & 53.86 & 86.76      & \textbf{95.87} & 29.56 & 95.82	  & 87.85 & 48.69 & 87.27 \\ \hline

$\hat{\theta} - \lambda_{tgt}\tau_{tgt}^{sa}$ NTK        & 84.43 & 73.71 & 90.24    & 81.78 & 20.00 & 85.86       & 88.41 & 76.92 & 91.85    & 83.26 & 44.47 & 86.98      & 93.48 & 53.12 & 94.96	  & 85.61 & 2.53 & 86.31 \\ \hline

\end{tabular}
}
\end{table*}

\begin{table*}[t]\renewcommand{\arraystretch}{1.3}
\caption{Accuracy ($\downarrow$) on the training set of the unlearned client at the end of the first FU round (FU) and at the end of the whole pipeline (PU).}
\label{model_target_acc_ablation}
\centering
\setlength{\tabcolsep}{0.4em}
\resizebox{\textwidth}{!}{%
\begin{tabular}{l@{\hskip 15pt}|*{2}{c}|*{2}{c}|*{2}{c}|*{2}{c}|*{2}{c}|*{2}{c}}
\hline
\multirow{3}{*}{Method}         & \multicolumn{4}{c}{$\beta$ = 0.05} & \multicolumn{4}{c}{$\beta$ = 0.1} & \multicolumn{4}{c}{$\beta$ = 0.5} \\ \cline{2-13}
                                & \multicolumn{2}{c}{Resisc45} & \multicolumn{2}{c}{Cars196} & \multicolumn{2}{c}{Resisc45} & \multicolumn{2}{c}{Cars196} & \multicolumn{2}{c}{Resisc45} & \multicolumn{2}{c}{Cars196} \\ \cline{2-13} 
                                & FU & PU & FU & PU & FU & PU & FU & PU & FU & PU & FU & PU \\ \hline

$\theta_0 + \sum_{i \neq tgt}\lambda_i \tau_i^{sa}$            & 58.91 & 62.08               & 53.73 & 47.69            & 81.85 & 90.46                & 55.63 & 56.73    	      & 91.76 & 94.03                   & 74.04 & 82.37    \\ \hline

$\theta_0 + \sum_{i \neq tgt}\lambda_i \tau_i^{sa}$ NTK        & 54.85 & 61.20               & 52.54 & 53.49            & 79.69 & 85.85                & 56.92 & 54.65    	      & 86.55 & 90.93                   & 71.03 & 77.93    \\ \hline

$\hat{\theta} - \lambda_{tgt}\tau_{tgt}^{sa}$            & 55.56 & 64.37               & 7.22 & \textbf{45.68}            & \textbf{31.08} & 85.03                & 10.53 & \textbf{51.29}    	      & 77.1 & 91.69                   & 26.18 & 77.88    \\ \hline

$\hat{\theta} - \lambda_{tgt}\tau_{tgt}^{sa}$ NTK      & \textbf{9.96} & \textbf{53.70}               & \textbf{1.48} & 49.17           & 31.69 & \textbf{81.44}                 & \textbf{6.36} & 57.59   	      & \textbf{14.29} & \textbf{90.48}                  & \textbf{0.27} & \textbf{74.15}    \\ \hline
\end{tabular}
}
\end{table*}

\begin{figure}[t]
    \centering

    \begin{subfigure}[t]{0.48\textwidth}
        \includegraphics[width=\linewidth]{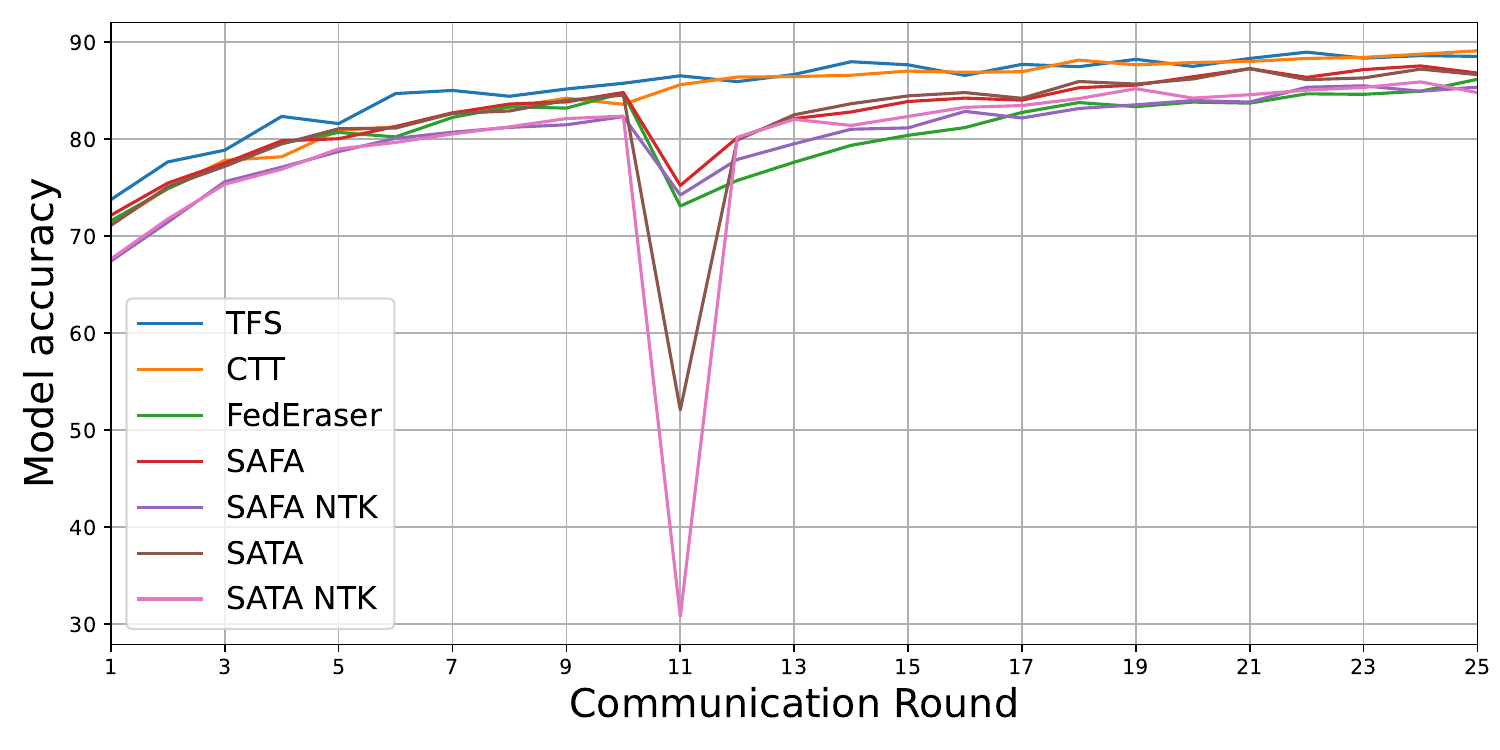}
        \caption{Accuracy on test set. $\beta = 0.05$}
    \end{subfigure}
    \hfill
    \begin{subfigure}[t]{0.48\textwidth}
        \includegraphics[width=\linewidth]{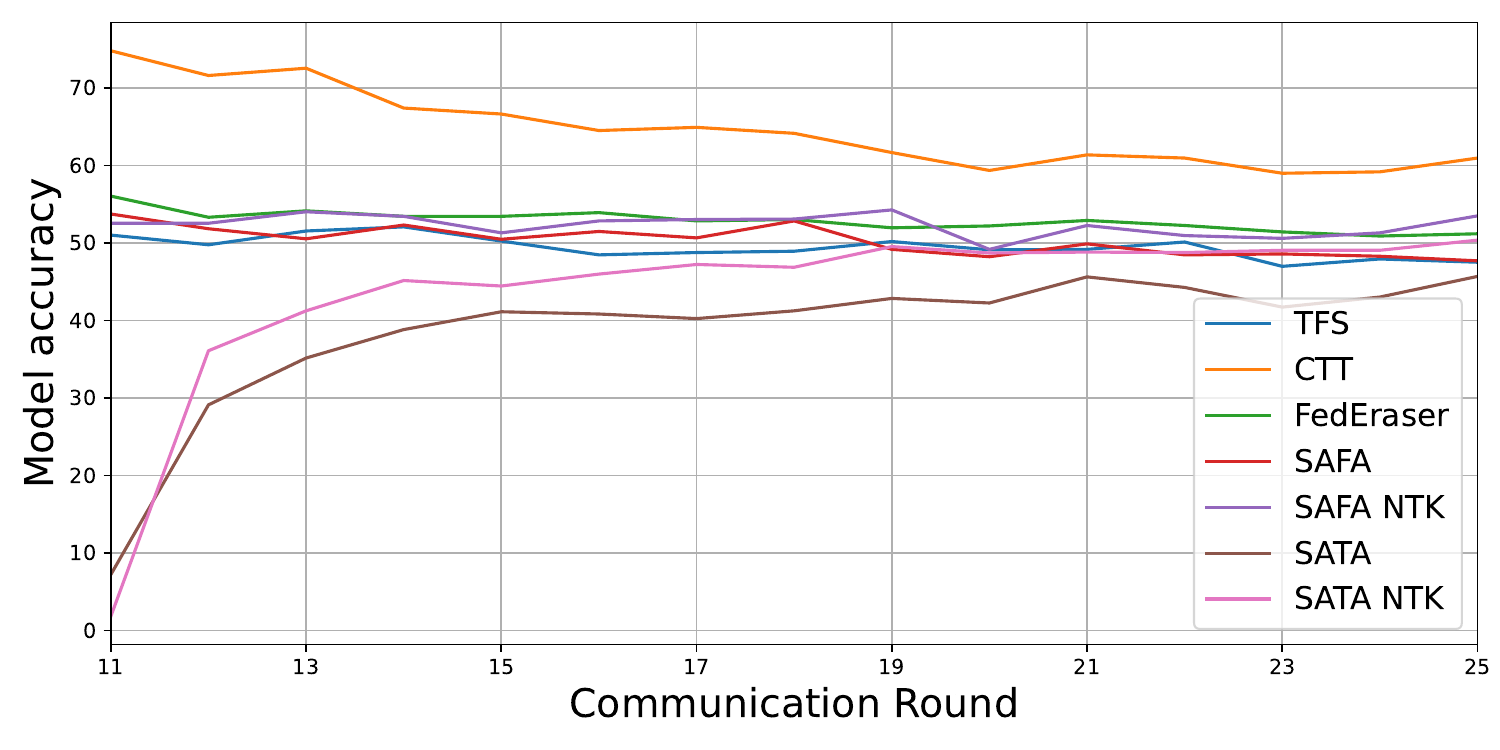}
        \caption{Accuracy on target testing set. $\beta = 0.05$}
    \end{subfigure}

    \vspace{0.5em} %

    \begin{subfigure}[t]{0.48\textwidth}
        \includegraphics[width=\linewidth]{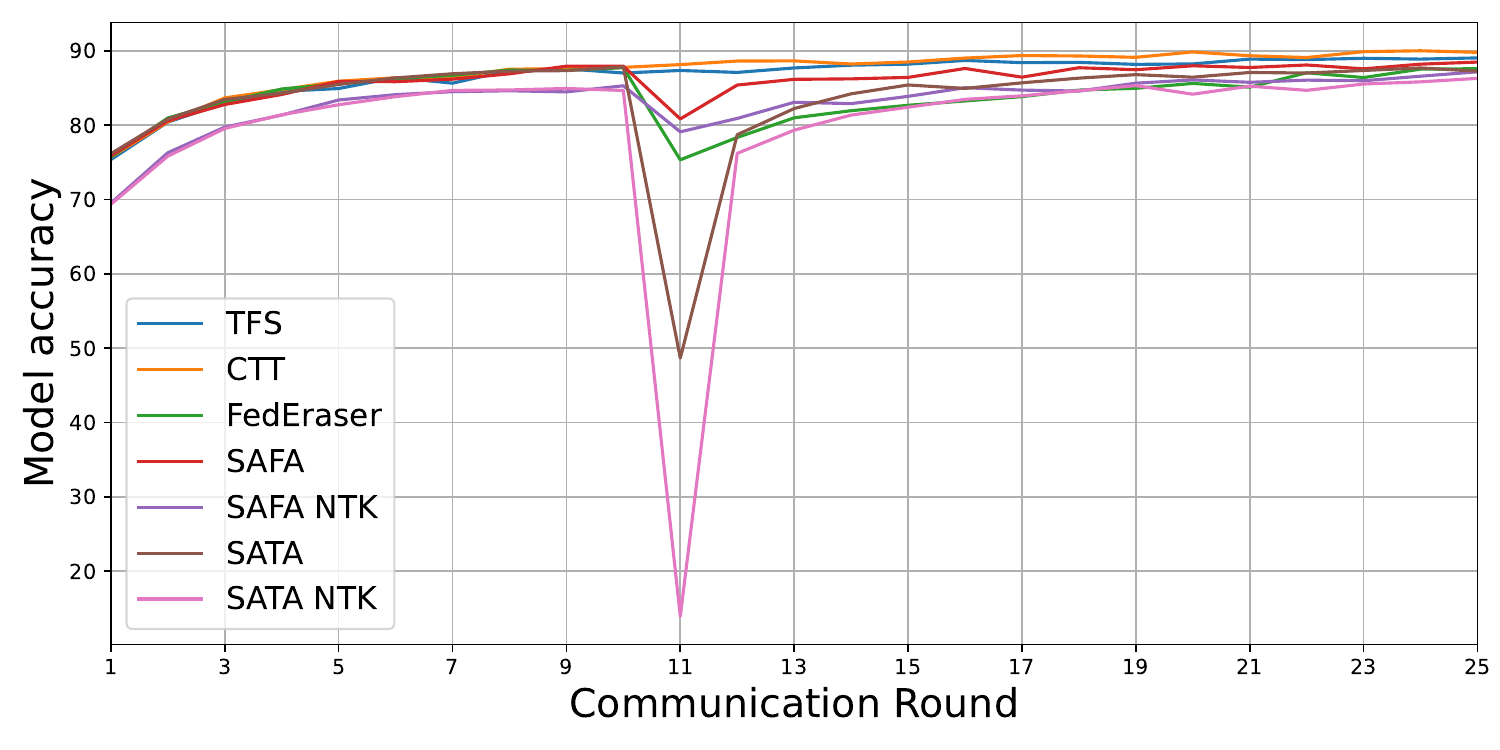}
        \caption{Accuracy on test set. $\beta = 0.5$}
    \end{subfigure}
    \hfill
    \begin{subfigure}[t]{0.48\textwidth}
        \includegraphics[width=\linewidth]{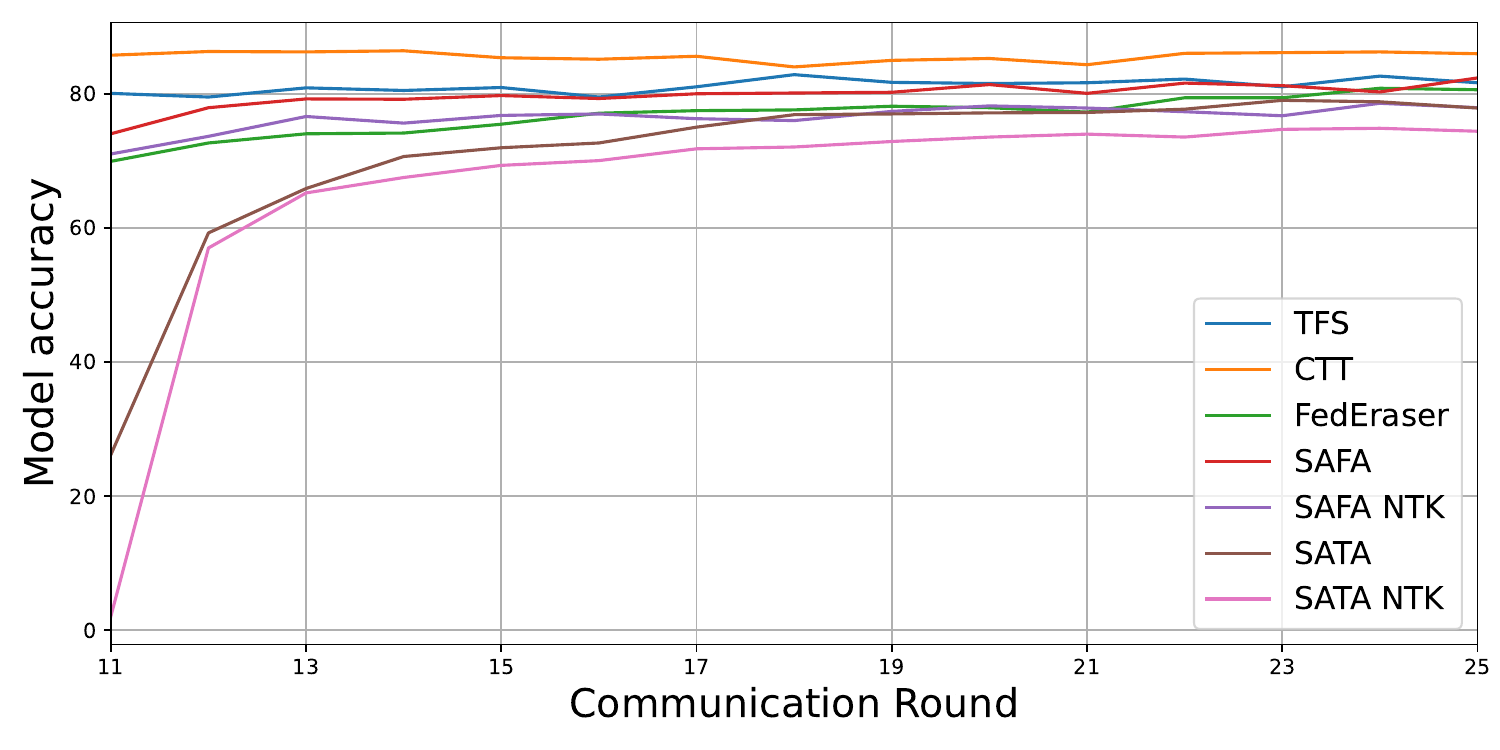}
        \caption{Accuracy on target testing set. $\beta = 0.5$}
    \end{subfigure}

    \caption{Plots of model accuracy on the Cars196 dataset.}
    \label{plots}
\end{figure}

\noindent
As reported in \cref{model_target_acc}, and \cref{plots}, our method successfully unlearns the target outperforming all competitors across every phase of the experiments. However, from \cref{model_acc} we observe that SATA leads the model to converge to slightly lower accuracies. In \cref{model_acc_ablation} and \cref{model_target_acc_ablation}, we present a comparison between our method in the NTK regime and its counterpart outside of it. The NTK regime consistently proves advantageous, improving unlearning performance regardless of the Dirichlet coefficient. We include an additional experiment, tested both in and out of the NTK regime, which applies FedAvg on the $\tau^{sa}$ of the remaining clients to quickly produce a new cleansed model in a single round. We refer to this approach as SAFA (Stand Alone FedAvg). Compared to SATA, the server model attains higher accuracy, though its unlearning performance is slightly weaker. Both SATA and SAFA reach convergence within only a few PU rounds.

\section{Conclusions}

In this work, we proposed a novel Federated Unlearning method that effectively removes a client's influence from a global model in just one communication round. Our approach, based on Task Arithmetic and enhanced through Neural Tangent Kernel training, enables fast and efficient unlearning while maintaining model utility. We showed that our method achieves competitive performance compared to existing unlearning approaches. Unlike traditional methods such as FedEraser, which require multiple communication rounds, our approach minimizes disruption to the federated system and allows for quick model re-usability. Future work will focus on addressing the method’s limitations under high Dirichlet coefficients and exploring its adaptability to other modalities and federated settings.

\bibliographystyle{splncs04}
\bibliography{main}
\end{document}